\documentclass[runningheads]{llncs}
\usepackage{graphicx}
\usepackage{multirow}
\usepackage{amsmath}
\usepackage{amsfonts}
\usepackage{tikz}
\usepackage{algorithm}
\usepackage{algpseudocode}
\usepackage{hyperref}
\def\checkmark{\tikz\fill[scale=0.4](0,.35) -- (.25,0) -- (1,.7) -- (.25,.15) -- cycle;}
\begin{document}
\title{Label-Efficient 3D Brain Segmentation via Complementary 2D Diffusion Models with Orthogonal Views}
\titlerunning{Label-Efficient 3D Brain Segmentation via 2D Diffusion Models}
%

\author{Jihoon Cho\inst{1} \and
Suhyun Ahn\inst{1} \and
Beomju Kim\inst{1} \and
Hyungjoon Bae\inst{2} \and
Xiaofeng Liu\inst{3} \and
Fangxu Xing\inst{4} \and
Kyungeun Lee\inst{2} \and
Georges Elfakhri\inst{3} \and
Van Wedeen\inst{4} \and
Jonghye Woo\inst{4} \and
Jinah Park\inst{1}
}

%
\authorrunning{Jihoon Cho et al.}
%


\institute{School of Computing, Korea Advanced Institute of Science and Technology, Daejeon 34141, Republic of Korea \\
\email{\{zinic,jinahpark\}@kaist.ac.kr} \and 
Electrical Engineering and Computer Science, Daegu Gyeongbuk Institute of Science and Technology, Daegu 42988, Republic of Korea \and
School of Medicine, Yale University, New Haven, CT 06510, USA \and
Gordon Center for Medical Imaging, Massachusetts General Hospital and Harvard Medical School, Boston, MA 02114, USA
}

%
\maketitle              
\begin{abstract}
Deep learning-based segmentation techniques have shown remarkable performance in brain segmentation, yet their success hinges on the availability of extensive labeled training data. Acquiring such vast datasets, however, poses a significant challenge in many clinical applications. To address this issue, in this work, we propose a novel 3D brain segmentation approach using complementary 2D diffusion models. The core idea behind our approach is to first mine 2D features with semantic information extracted from the 2D diffusion models by taking orthogonal views as input, followed by fusing them into a 3D contextual feature representation. Then, we use these aggregated features to train multi-layer perceptrons to classify the segmentation labels. Our goal is to achieve reliable segmentation quality without requiring complete labels for each individual subject. Our experiments on training in brain subcortical structure segmentation with a dataset from only one subject demonstrate that our approach outperforms state-of-the-art self-supervised learning methods. Further experiments on the minimum requirement of annotation by sparse labeling yield promising results even with only nine slices and a labeled background region. 


\keywords{Segmentation \and Diffusion model \and Brain MRI.}
\end{abstract}
\section{Introduction}

Brain MRI segmentation has become a crucial step in numerous clinical studies, as it significantly influences the overall outcomes of the analysis, including diagnosis, disease progression tracking, and treatment monitoring~\cite{despotovic2015mri}. Yet, manual segmentation of various brain structures is a time-consuming and labor-intensive process. Over the past several years, deep learning-based segmentation methods have shown accurate prediction results with volume-based processing, which inherits 3D anatomical information~\cite{cao2022swin,cho2022hybrid,hatamizadeh2021swin,isensee2021nnu}. However, producing reliable segmentation results requires a substantial number of training labels, which leads to a contradictory situation where additional manual labels are required to solve the problem of difficulties in manual segmentation. As a result, new clinical analyses, which struggle to benefit from automatic segmentation methods due to insufficient labeled data, still rely on labor-intensive manual segmentation.

To address this issue, self-supervised learning (SSL) has demonstrated improved data efficiency by generating useful feature representations using available unlabeled data. The capability of solving the pretext task through an SSL approach enables effective initialization of network weights for subsequent downstream tasks~\cite{jin2023labelefficient}. In particular, understanding the 3D anatomical structure, when using SSL, plays a vital role in achieving accurate segmentation results, even with only a few limited labels~\cite{zhou2021models}. As such, an image synthesis method, taking into account local and global structural details without relying on labeled data, can also serve as an effective representation learner for semantic segmentation~\cite{tritrong2021repurposing,xu2021linear,zhang2021datasetgan}. Especially, recently proposed denoising diffusion probabilistic models (DDPMs)~\cite{dhariwal2021diffusion,ho2020denoising} have shown impressive segmentation performance via a gradual denoising process, by creating features of fine-grained semantic information from specific diffusion timesteps~\cite{baranchuk2021label,yang2023diffusion}, although they have been limited to 2D natural images.

Leveraging the benefits of diffusion models, in this work, we introduce a novel approach to 3D medical image segmentation. To mitigate the significant computational expenses associated with the 3D generative approach, our proposed architecture relies on perpendicular and complementary 2D diffusion models, taking orthogonal orientations as input. Our 2D diffusion models conduct denoising with orthogonal views from a volume to extract semantic information, which is then integrated into 3D representations to enhance the level of information provided. Our approach, which employs simple MLPs for classifying segmentation labels, has demonstrated superior results over state-of-the-art (SOTA) self-supervised segmentation methods in subcortical brain structure segmentation using only one labeled dataset. In our further analysis, we focus on the effectiveness of the labeling scheme based on the MLPs. Our approach, which incorporates a sparse labeling scheme, has shown that promising 3D segmentation results can be achieved with just nine labeled slices and an easily identifiable background region.

Our main contributions can be summarized as follows:
\begin{itemize}
    \item We propose a novel and efficient 3D segmentation approach that leverages the extraction of 3D semantic information from complementary 2D diffusion models with orthogonal orientations as input.
    \item We demonstrated, through successful segmentation of the brain subcortical structure, that even with only one labeled training dataset, a simple MLP can achieve precise predictions by aggregating complementary information.
    \item Our experimental results demonstrated that with sparse annotation, consisting of only nine slices and a background region, the performance of our proposed approach surpassed the SOTA SSL methods.
\end{itemize}

\begin{figure}[t]
\begin{center}
\includegraphics[width=1.0\linewidth]{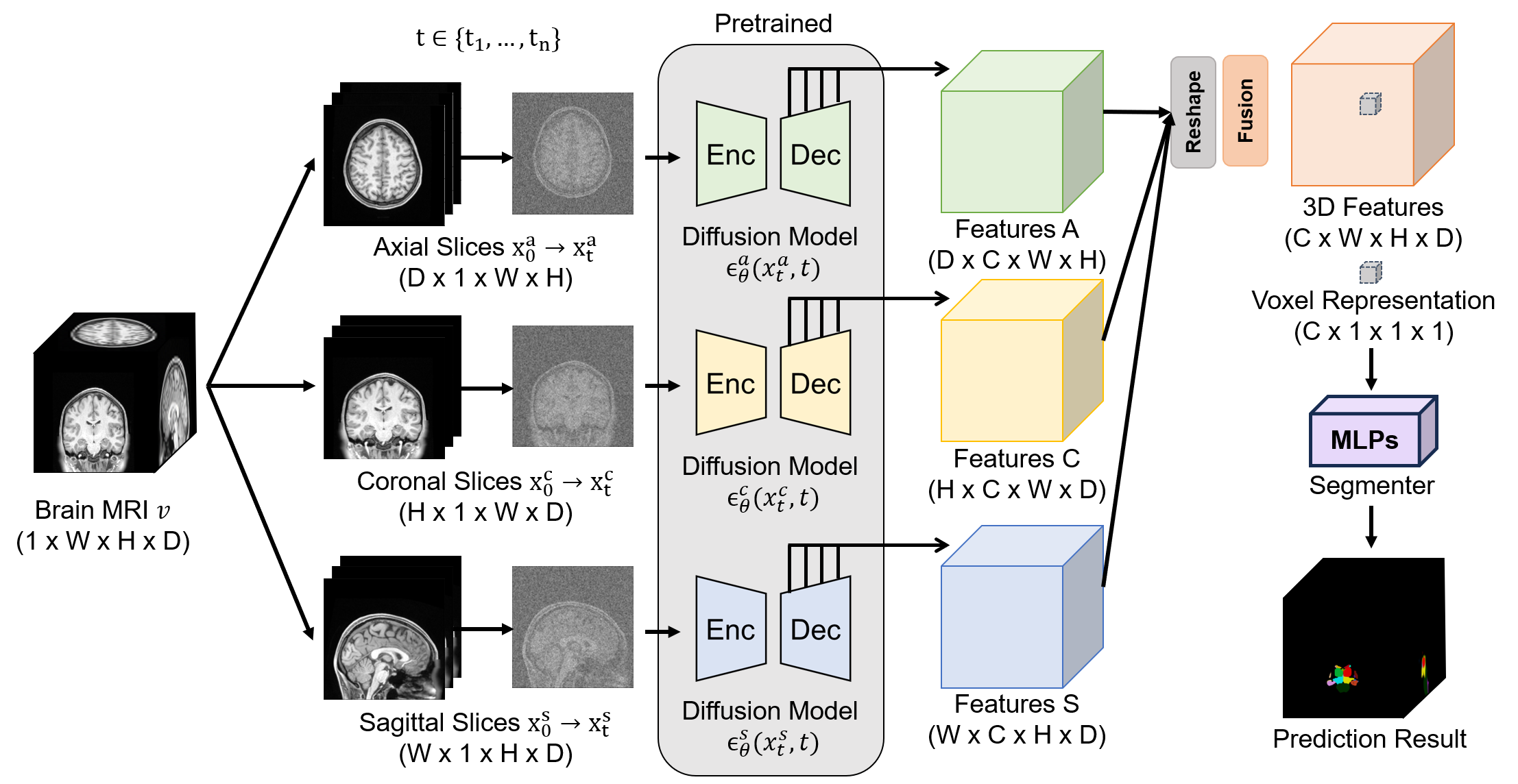}
\end{center} 
\caption{Illustration of our proposed method on subcortical brain structure segmentation. We first train multiple 2D diffusion models using an enormous unlabeled dataset. MLPs are then trained with a few labels and predict segmentation results by leveraging 3D features transformed from three perpendicular 2D diffusion models.}
\label{fig:overview}
\end{figure} 

\section{Method}
In this section, we provide a brief overview of the DDPM and relevant research on semantic information. Then, we describe the process of establishing a 3D environment for segmentation using individual pre-trained 2D diffusion models. We also present the sparse labeling scheme designed for efficient training using a simple MLP segmenter. An overview of the proposed method is illustrated in~\autoref{fig:overview}.

\subsection{Preliminaries}
Diffusion models are a class of deep generative models that generate data $x_0$ sampled from a data distribution by gradually denoising noise $x_T$ from a simple distribution to a less noisy sample $x_t$. To model a data distribution, DDPMs~\cite{ho2020denoising} obtain a noisy sample $x_t$ directly from the data $x_0$ using a forward process, which is gradually adding Gaussian noise based on a variance schedule $\beta_1,...,\beta_T$, and approximate a reverse process:
\begin{equation}
    q(x_t|x_0):=N(x_t; \sqrt{\bar{\alpha_t}}x_0, (1-\bar{\alpha_t})I),
\label{eq:forward}
\end{equation}
\begin{equation}
    p_\theta(x_{t-1}|x_t):=N(x_{t-1}; \mu_\theta(x_t,t), \mathrm{\Sigma}_\theta(x_t,t)),
\label{eq:backward}
\end{equation}
where $\alpha_t := 1 - \beta_t$ and $\bar{\alpha_t} :=$$ \mathrm{\Pi}_{s=1}^{t}\alpha_s$.
 
\noindent A noise predictor $\epsilon_\theta(x_t,t)$ takes a noisy input $x_t$ and the timestep $t$ as input and predicts a simplified noise $\tilde{\epsilon}$ rather than predicts $\tilde{\mu_t}$ as $\mu_\theta$. U-Net architecture~\cite{ronneberger2015u} is used as a noise predictor, and DDPM-Seg~\cite{baranchuk2021label} has determined that the feature representation of the network provides semantic information from an input image. In particular, when dealing with an image with less noise, such as a small $t$, the U-Net decoder blocks effectively capture semantic information from coarse to fine-grained details through progressive decoding stages.

\subsection{3D Semantic Information Extraction from 2D Diffusion Models}
Accurately segmenting 3D images requires acquiring contextual information about the target structure. Due to the substantial computational demands of 3D diffusion models, the use of 2D diffusion models becomes necessary to extract semantic information. However, 2D diffusion models handle image slices independently, resulting in a lack of interaction between neighboring slices.

To address this issue, we employ multiple 2D diffusion models trained using image slices from orthogonal orientations, including axial, coronal, and sagittal views. When a 3D volume $v \in \mathbb{R}^{1\times W\times H\times D}$ is provided, we partition the volume $v$ along various axes and use these slices, which are axial slices $x_0^{a} \in \mathbb{R}^{D\times 1\times W\times H}$, coronal slices $x_0^{c} \in \mathbb{R}^{H\times 1\times W\times D}$, and sagittal slices $x_0^{s} \in \mathbb{R}^{W\times 1\times H\times D}$, as inputs after corrupting $x_0$ by adding Gaussian noise as described in~\autoref{eq:forward}. The predefined compositions of the intermediate activation, extracted from the U-Net decoder blocks $B$ and the diffusion steps $t$, are upsampled to the shape of the input $x_t$ using bilinear interpolation. Subsequently, we reshape the extracted features to match the original shape of the 3D volume $v$ and combine all three sets of features from different diffusion models. Thus, it allows us to extend the 2D information from each view into voxel-level representations of $v$.

\begin{figure}[t]
\begin{center}
\includegraphics[width=1.0\linewidth]{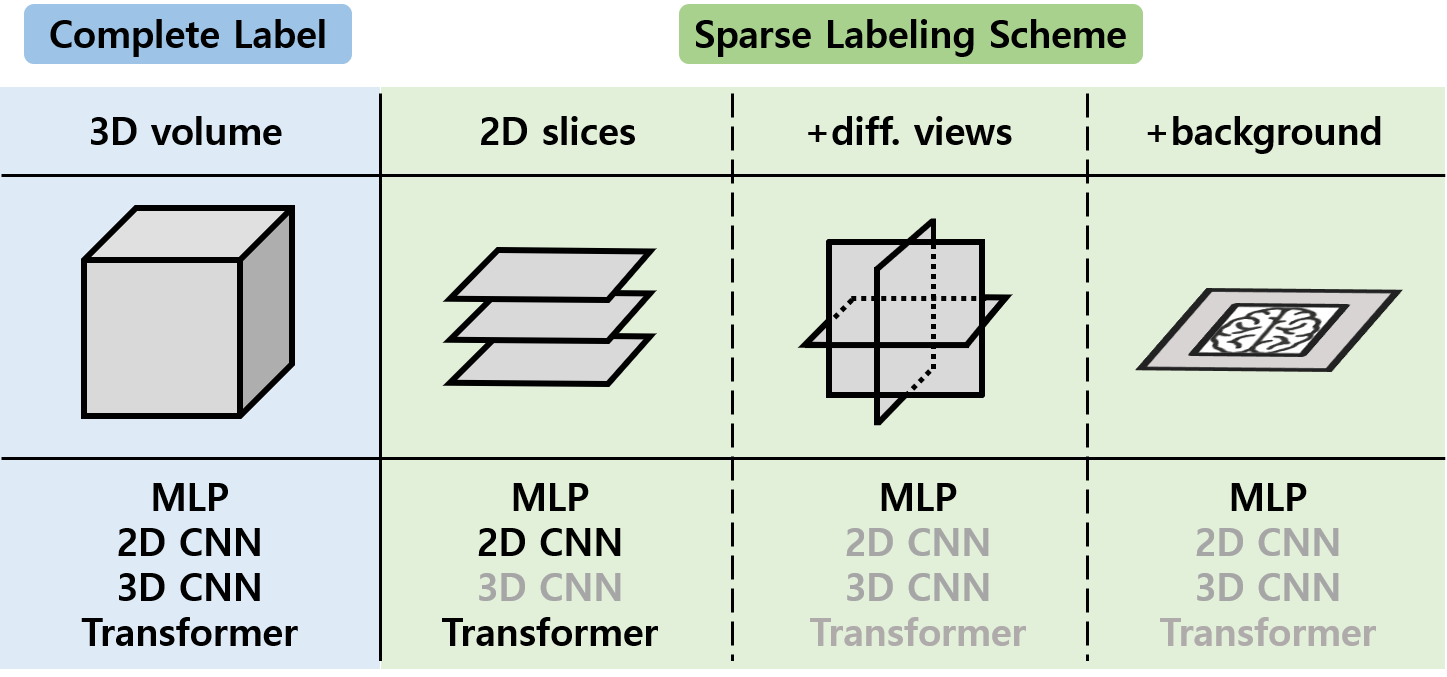}
\end{center} 
\caption{The schematic diagram of the proposed sparse labeling scheme. A complete label requires the annotation of entire voxels. The sparse labeling scheme is based on 2D labeled slices annotated from different views with an easily identifiable background region. The last row represents the segmenters that can be trained for each case of label.}
\label{fig:labeling}
\end{figure} 


\subsection{Simple MLP Segmenter and Sparse Labeling Scheme}
The extracted feature representations from the proposed approach are utilized to train a segmenter that predicts semantic labels for every voxel. These features have already integrated 3D semantic information and do not require complex processing. Hence, a simple MLP is sufficient to classify the segmentation labels. Another benefit of MLP is its ability to process each voxel independently. Therefore, our proposed approach using MLP can consistently capture 3D anatomical details regardless of the shapes of the segmentation labels. In contrast, SSL methods with 2D networks face a limitation in the lack of 3D anatomical information.

In addition, we introduce a sparse labeling scheme to achieve promising segmentation results while minimizing the need for manual labeling. An example of a sparse labeling scheme is shown in~\autoref{fig:labeling}. The key concept is to reduce the necessary annotated slices by strategically selecting them. Since our approach does not depend on a particular orientation, we employ various label slices of orthogonal views. It allows a broader range of 3D anatomical structures to be contained, regardless of the location, direction, or shape of the target organ. We also take into account the presence of easily determinable background slices that do not contain target organs, aiming to achieve robust predictions by minimizing misclassification. This can be accomplished by identifying just two slices from the final slices of the target structures for each axis.

\section{Experiments}
\subsection{Experimental Settings}
We conducted our experiments on the dataset of the human connectome project (HCP)~\cite{van2013wu}, from which we used unlabeled T1-weighted MRI scans. A total of 1,015 unlabeled volumes were used for pre-training, and a total of 12 volumes were used for the downstream segmentation task with manual segmentation of 28 brain subcortical structures from the Harvard-Oxford Atlas 2.0 project~\cite{rushmore2022anatomically}. The 12 volumes for the segmentation task were divided into one for training, one for validation, and 10 for testing. All volumes were normalized to $[-1,1]$ with the 99.5\% percentile of intensity and reshaped to $256\times 256\times 256$, while maintaining isotropic resolution with zero padding.

For a fair comparison, we follow the same settings as in DDPM-Seg~\cite{baranchuk2021label}, which are the pre-trained diffusion model~\cite{dhariwal2021diffusion} consisting of 18 decoder blocks, diffusion timestep $T=1000$, and the composition of the features extracted from $B=\{5,6,7,8,12\}$ and $t=\{50, 150, 250\}$. We train the 2D diffusion models for 100,000 iterations using unlabeled volumes, which corresponds to a total of 259,840 slices. We performed a comparison of three fusion methods: selecting the maximum value among three different features (max), concatenating all features together (concat), and summing up the values of the three features (sum). Our segmenter has a simple structure of two hidden layers with the ReLU activation function and batch normalization, following the MLP architecture of DatasetGAN~\cite{zhang2021datasetgan}. We trained ten independent segmenters using cross-entropy loss for at least 4 epochs until convergence. To enhance the robustness of the predictions, we applied an ensemble method to ten prediction results. Additionally, we applied post-processing using the connected component algorithm and specifically selected the largest component to predict the brainstem. This approach addresses the issue of class imbalance, given that the brainstem is the largest among the 28 subcortical brain structures. 

\begin{table*}[t]
\begin{center}
\resizebox{1.0\textwidth}{!}{
\begin{tabular}{l|c|c|cc|cc}
\hline
\multirow{2}{*}{Methods} & \multirow{2}{*}{Input} & \multirow{2}{*}{Segmenter} & \multicolumn{2}{c|}{Dice Score (\%) $\uparrow$} & \multicolumn{2}{c}{HD95 (mm) $\downarrow$} \\ \cline{4-7} 
 & & &  {Avg.} & {Avg\dag} & {Avg.} & {Avg\dag} \\
\hline
3D U-Net~\cite{ronneberger2015u}        & 3D volume & 3D CNN & 73.91 & 76.14 & 8.00 & 5.41 \\
+Denoising~\cite{eun2020deep}           & 3D volume & 3D CNN & 73.10 & 75.22 & \textbf{6.46} & 2.46 \\
+Models Genesis~\cite{zhou2021models}    & 3D volume & 3D CNN & 75.44 & 78.21 & 6.57 & 2.17 \\
\hline
2D U-Net~\cite{ronneberger2015u}        & 2D slices (axial) & 2D CNN & 62.75 & 65.07 & 45.00 & 36.37 \\
+Jigsaw~\cite{noroozi2016unsupervised}  & 2D slices (axial) & 2D CNN & 56.86 & 58.97 & 46.32 & 46.35 \\
+RPL~\cite{doersch2015unsupervised}     & 2D slices (axial) & 2D CNN & 59.63 & 61.82 & 27.37 & 23.29 \\
+Denoising\cite{eun2020deep}            & 2D slices (axial) & 2D CNN & 69.29 & 71.86 & 43.60 & 39.32 \\
+Models Genesis~\cite{zhou2021models}    & 2D slices (axial) & 2D CNN & 67.93 & 70.45 & 34.83 & 22.16 \\
\hline
MAE~\cite{he2022masked}                 & 2D slices (axial) & Transformer & 23.65 & 24.52 & 30.87 & 18.05 \\
\hline
DDPM-Seg~\cite{baranchuk2021label}      & 2D slices (axial) & MLP & 74.24 & 76.99 & 15.53 & 2.14 \\
Ours (max)          & 2D slices (3 views) & MLP & 74.33 & 77.08 & 20.83 & 7.63 \\
Ours (concat)       & 2D slices (3 views) & MLP & 76.48 & 79.32 & 16.70 & 3.35 \\
Ours (sum)          & 2D slices (3 views) & MLP & \textbf{77.32} & \textbf{80.19} & 15.17 & \textbf{1.76} \\
\hline
\end{tabular}
}
\end{center}
\caption{Results on subcortical brain structure segmentation. The networks were pre-trained using an unlabeled dataset and then fine-tuned with a single volume. The 2D U-Net and 3D U-Net~\cite{ronneberger2015u} were only trained with a single volume without undergoing a pre-training stage.}
\label{table:result1}
\end{table*}

\subsection{Quantitative Comparisons}
We evaluated our approach and the SOTA SSL methods. We used 2D and 3D U-Net~\cite{ronneberger2015u} as baseline networks for CNN-based segmenter and employed Jigsaw~\cite{noroozi2016unsupervised}, RPL~\cite{doersch2015unsupervised}, Denoising~\cite{eun2020deep}, and Models Genesis~\cite{zhou2021models} as pre-trained SSL methods. To assess the performance of a segmenter based on transformers, we used MAE~\cite{he2022masked}. The models were trained for 300,000 iterations for 2D networks and at least 100,000 iterations for 3D networks, following the same settings as in the prior work. Fine-tuning was then carried out on the networks for the segmentation task using dice loss and cross-entropy loss until convergence was achieved. We used the Dice score and the Hausdorff distance with the 95\% percentile (HD95) as evaluation metrics. If no label prediction is present, we impose a penalty on HD95 by assigning a maximum distance of 377.07mm. Moreover, the structure of the fifth ventricle and the optic chiasm were represented by only nine and 56 voxels in the training data, respectively. So, we additionally reported analysis results, excluding these structures. The symbols Avg., Avg\dag, and Avg\ddag\ denote the results calculated for 28 subcortical brain structures, 27 structures omitting the fifth ventricle, and 26 structures excluding both the fifth ventricle and the optic chiasm, respectively.

\begin{table*}[t]
\begin{center}
\resizebox{1.0\textwidth}{!}{
\begin{tabular}{l|ccc|ccc|ccc}
\hline
\multirow{2}{*}{Methods} & \multicolumn{3}{c|}{Training Data} & \multicolumn{3}{c|}{Dice Score (\%) $\uparrow$} & \multicolumn{3}{c}{HD95 (mm) $\downarrow$} \\ \cline{2-10} 
 & axial & coronal & sagittal & {Avg.} & {Avg\dag} & {Avg\ddag} & {Avg.} & {Avg\dag} & {Avg\ddag} \\
\hline
DDPM-Seg~\cite{baranchuk2021label}  & \checkmark & & & 74.24 & 76.99 & 79.70 & 15.53 & 2.14 & 1.96 \\
DDPM-Seg~\cite{baranchuk2021label}  & & \checkmark & & 74.05 & 76.79 & 79.74 & 27.71 & 14.77 & 2.25 \\
DDPM-Seg~\cite{baranchuk2021label}  & & & \checkmark & 71.95 & 74.61 & 76.96 & 17.71 & 4.40 & 4.38 \\
\hline
Ours                    & \checkmark & \checkmark & & 76.48 & 79.31 & 81.64 & 15.31 & 1.91 & 1.79 \\
Ours                    & \checkmark & & \checkmark & 76.33 & 79.15 & 81.71 & 15.26 & 1.86 & 1.76 \\
Ours                    & & \checkmark & \checkmark & 76.16 & 78.98 & 81.65 & 15.41 & 2.02 & 1.87\\
\hline
Ours                    & \checkmark & \checkmark & \checkmark & \textbf{77.32} & \textbf{80.19} & \textbf{82.35} & \textbf{15.17} & \textbf{1.76} & \textbf{1.70} \\
\hline
\end{tabular}
}
\end{center}
\caption{Ablation study using a combination of different feature representations from different views. The MLP segmenters were trained with a single volume.}
\label{table:result2}
\end{table*}

\begin{table*}[b]
\begin{center}
\resizebox{1.0\textwidth}{!}{
\begin{tabular}{l|ccc|ccccc}
\hline
\multirow{2}{*}{Methods} & \multicolumn{3}{c|}{Training Labels} & \multicolumn{5}{c|}{Number of Slices} \\ 
\cline{2-9} 
 & axial & 3-views & bg & {9} & {10} & {15} & {30} & {50} \\
\hline
2D UNet~\cite{ronneberger2015u}   & \checkmark & & & - & 29.44(142.41) & - & 43.43(162.80) & 51.30(147.45) \\
+Jigsaw~\cite{noroozi2016unsupervised}   & \checkmark & & & - & 29.17(158.38) & - & 44.83(136.87) & 56.61(109.40) \\
+RPL~\cite{doersch2015unsupervised}      & \checkmark & & & - & 35.52(90.58) & - & 46.20(114.18) & 54.56(114.05) \\
+Denoising\cite{eun2020deep}             & \checkmark & & & - & 51.54(99.97) & - & 62.56(93.29) & 70.10(77.63) \\
+Models Genesis~\cite{zhou2021models}     & \checkmark & & & - & 39.07(132.56) & - & 52.21(118.06) & 57.49(113.85) \\
\hline
Ours & \checkmark & & & - & \textbf{56.95(59.98)} & - & 79.61(14.76) & \textbf{82.12(3.58)} \\
Ours & & \checkmark & & 65.36(41.49) & - & 80.52(5.06) & \textbf{81.91(1.76)} & - \\
Ours & & \checkmark  & \checkmark & \textbf{77.18(2.58)} & - & \textbf{80.53(1.96)} & 81.79(1.94) & - \\
\hline
\end{tabular}
}
\end{center}
\caption{Comparison results trained with different training configurations. The utilization of 3D networks is not feasible due to the incomplete labeling of a volume. The numbers written without brackets indicate Avg\ddag\ of the Dice score, and the numbers in brackets indicate Avg\ddag\ of the HD95.}
\label{table:result3}
\end{table*}

The comparison results are presented in~\autoref{table:result1}. Notable findings include that the transformer is not suitable for limited training data due to its lack of inductive bias, resulting in the worst MAE results. Conversely, generative SSL methods, such as Denoising and Models Genesis, have shown significant gains, leading to excellent results. In particular, the 3D network with Models Genesis demonstrated superior performance over DDPM-Seg, which lacks interactions between adjacent slices; however, our proposed approach outperformed all other SOTA methods with 80.19\% of the average Dice score and 1.76mm of the average HD95 by the construction of 3D context. The fusion method to add the values of the three features (sum) has been demonstrated to be the most advantageous method. This method provides a succinct abstraction that encompasses all feature representations, making it suitable for a simple MLP segmenter.
Furthermore, in the ablation study of the combination of multiple diffusion models, as you can see in~\autoref{table:result2}, our approach demonstrated the ability to produce 3D semantic information, by robust prediction results, even when combining only two perpendicular models.

\begin{figure}[t]
\begin{center}
\includegraphics[width=1.0\linewidth]{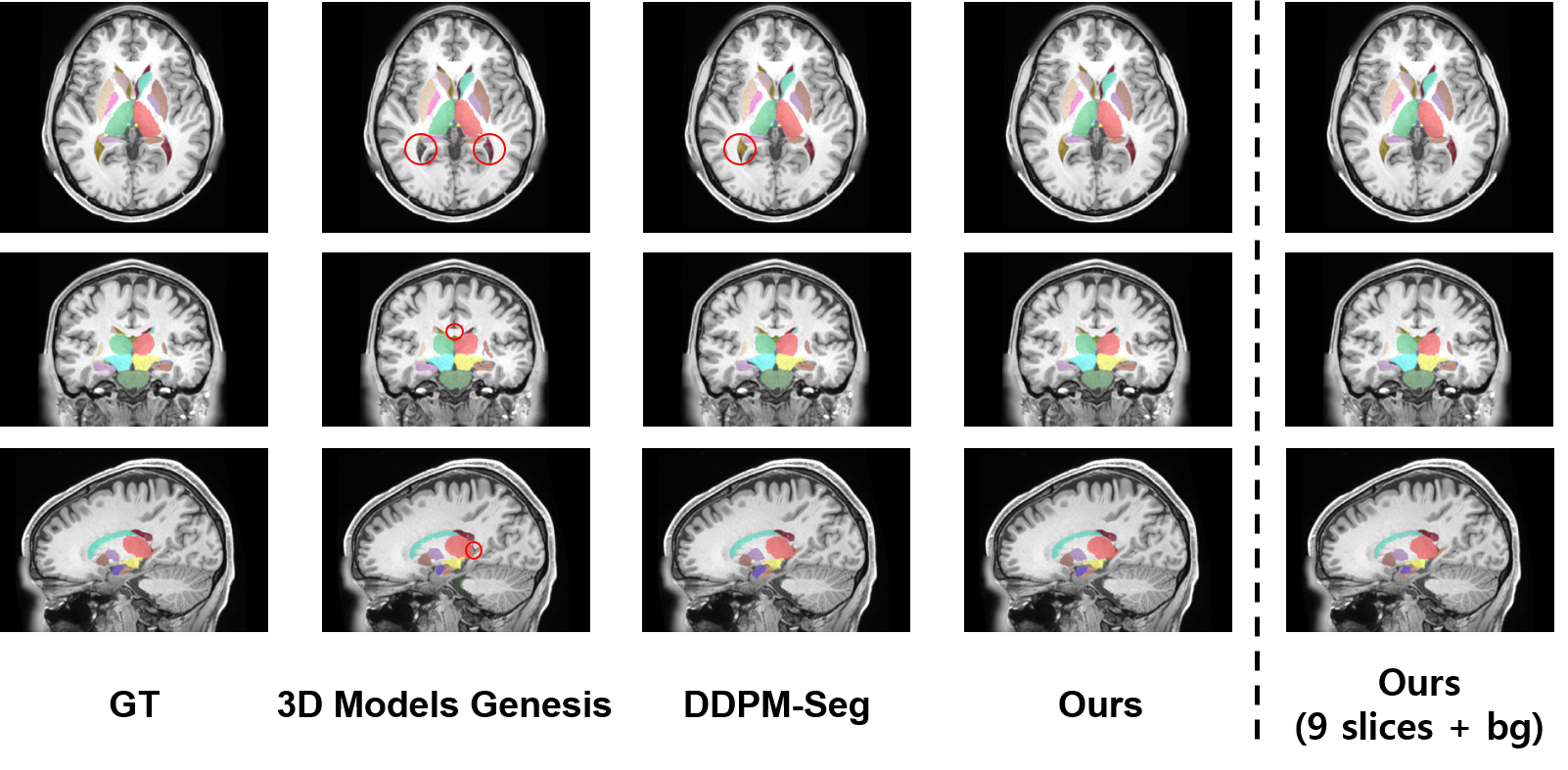}
\end{center} 
\caption{Segmentation results on subcortical brain structure segmentation. All of the segmentation results except the last column were predicted from the models trained with a single labeled volume. The mispredictions are indicated by the red circles.}
\label{fig:qualitative}
\end{figure}

\subsection{Experiments on Sparse Labeling Scheme}
In order to determine the minimal labeling required to yield promising results, we performed experiments with a few training slices. The training configurations consist of 3, 5, and 10 slices for each axis, and 10, 30, and 50 axial slices, with a stride of 10, 10, 5, 4, 2, and 2, respectively. As demonstrated in~\autoref{table:result3}, the prediction results from the SSL methods did not show encouraging results. In contrast, our approach has shown promising segmentation results with 50 axial slices. Moreover, by using an effective sparse labeling scheme that capitalizes on the benefits of our approach, we have achieved excellent performances of 77.18\% for the mean Dice score and 2.58mm for the mean HD95, for 26 brain subcortical structures, with only nine annotated slices and background regions. The inaccuracies in predicting the fifth ventricle and optic chiasm were due to a class imbalance issue, where the annotations were very small, consisting of only one and three voxels, respectively. The robust segmentation results obtained using our approach are shown in~\autoref{fig:qualitative}.

\section{Conclusion}
In this paper, we proposed a novel 3D segmentation approach to segment brain structures using a few labeled data. Specifically, we demonstrated our capacity to extract 3D information from 2D diffusion models and achieved promising results, even with very few slices being annotated. Our approach shows promise in the potential to reduce the required cost to build an automatic segmentation system. In our future work, we will explore methods for capturing semantic information in small regions, such as the fifth ventricle, which consists of only a few voxels in the training data.



%
%
%
%
\newpage

\bibliographystyle{splncs04}
\bibliography{MLMI}

\end{document}